\documentclass[sigconf]{acmart}
\AtBeginDocument{
  \providecommand\BibTeX{{
    \normalfont B\kern-0.5em{\scshape i\kern-0.25em b}\kern-0.8em\TeX}}}

\usepackage[linesnumbered,ruled]{algorithm2e}
\usepackage{balance}
\usepackage{booktabs}
\usepackage{enumitem}
\usepackage{epsfig}
\usepackage{gensymb}
\usepackage{graphics}
\usepackage{graphicx}
\usepackage{hhline}
\usepackage{listings}
\usepackage{multirow}
\usepackage{subcaption}
\usepackage{textcomp}
\usepackage{url}
\usepackage{wrapfig}
\usepackage{xspace}

\usepackage[nolist,nohyperlinks]{acronym}
\acrodef{AI}{Artificial Intelligence}
\acrodef{ANN}{artificial neural network}
\acrodef{ML}{Machine Learning}
\acrodef{SVM}{support vector machine}

\newcommand{\simulamet}{SimulaMet \country{Norway}}
\newcommand{\forzasys}{Forzasys AS \country{Norway}}
\newcommand{\oslomet}{Oslo Metropolitan University \country{Norway}}

\setcopyright{acmcopyright}
\copyrightyear{2022}
\acmYear{2022}
\acmDOI{}
\acmConference[MMSys'22]{13th ACM Multimedia
Systems Conference}{June 14-17, 2022}{Athlone, Ireland}
\acmBooktitle{13th ACM Multimedia Systems Conference (MMSys'22), June 14-17, 2022,
2022, Athlone, Ireland}
\acmPrice{15.00}
\acmISBN{}

\begin{document}

\title{MMSys'22 Grand Challenge on \\
AI-based Video Production for Soccer}

\author{Cise Midoglu}
\affiliation{
    \institution{\simulamet}
}
\author{Steven A. Hicks}
\affiliation{
    \institution{\simulamet}
}
\author{Vajira Thambawita}
\affiliation{
    \institution{\simulamet}
}
\author{Tomas Kupka}
\affiliation{
    \institution{\forzasys}
}
\author{Pål Halvorsen}
\affiliation{
    \institution{\simulamet}
}
\authornote{Also affiliated with \oslomet}
\authornote{Also affiliated with \forzasys}

\renewcommand{\shortauthors}{Midoglu et al.}

\begin{abstract}
Soccer has a considerable market share of the global sports industry, and the interest in viewing videos from soccer games continues to grow. In this respect, it is important to provide game summaries and highlights of the main game events. However, annotating and producing events and summaries often require expensive equipment and a lot of tedious, cumbersome, manual labor. Therefore, automating the video production pipeline providing fast game highlights at a much lower cost is seen as the ``holy grail''. In this context, recent developments in Artificial Intelligence (AI) technology have shown great potential. Still, state-of-the-art approaches are far from being adequate for practical scenarios that have demanding real-time requirements, as well as strict performance criteria (where at least the detection of official events such as goals and cards must be $100\%$ accurate). In addition, event detection should be thoroughly enhanced by annotation and classification, proper clipping, generating short descriptions, selecting appropriate thumbnails for highlight clips, and finally, combining the event highlights into an overall game summary, similar to what is commonly aired during sports news. Even though the event tagging operation has by far received the most attention, an end-to-end video production pipeline also includes various other operations which serve the overall purpose of automated soccer analysis. This challenge aims to assist the automation of such a production pipeline using AI. In particular, we focus on the enhancement operations that take place after an event has been detected, namely event clipping (Task 1), thumbnail selection (Task 2), and game summarization (Task 3). Challenge website: \url{https://mmsys2022.ie/authors/grand-challenge}.

\end{abstract}


\begin{teaserfigure}
\centering
    \begin{subfigure}[b]{0.33\textwidth}
      \includegraphics[width=\textwidth,height=3.5cm]{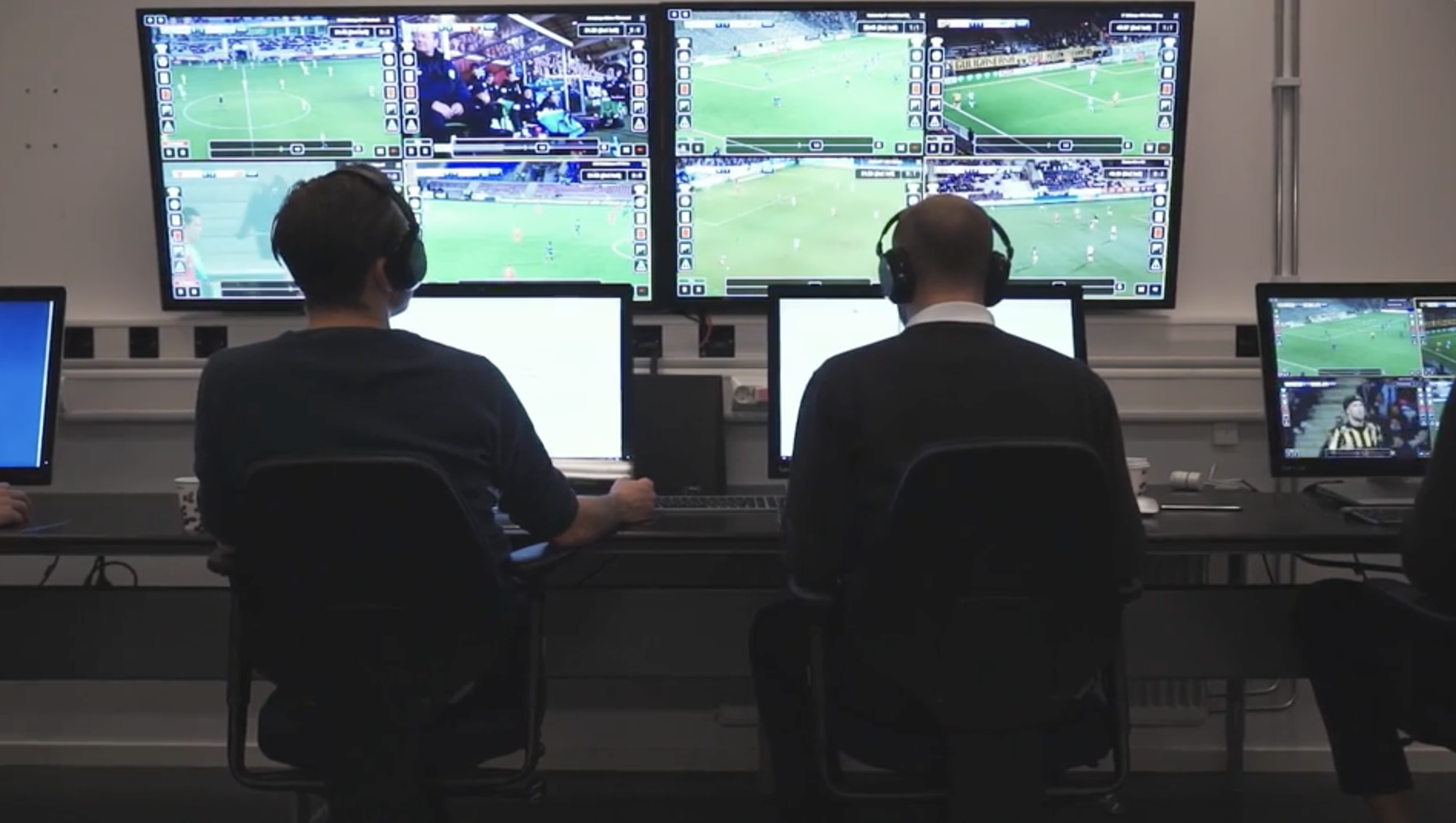}
      \caption{Manual operators at their workstation.}
      \label{figure:tagging1}
    \end{subfigure}
    \begin{subfigure}[b]{0.33\textwidth}
      \includegraphics[width=\textwidth,height=3.5cm]{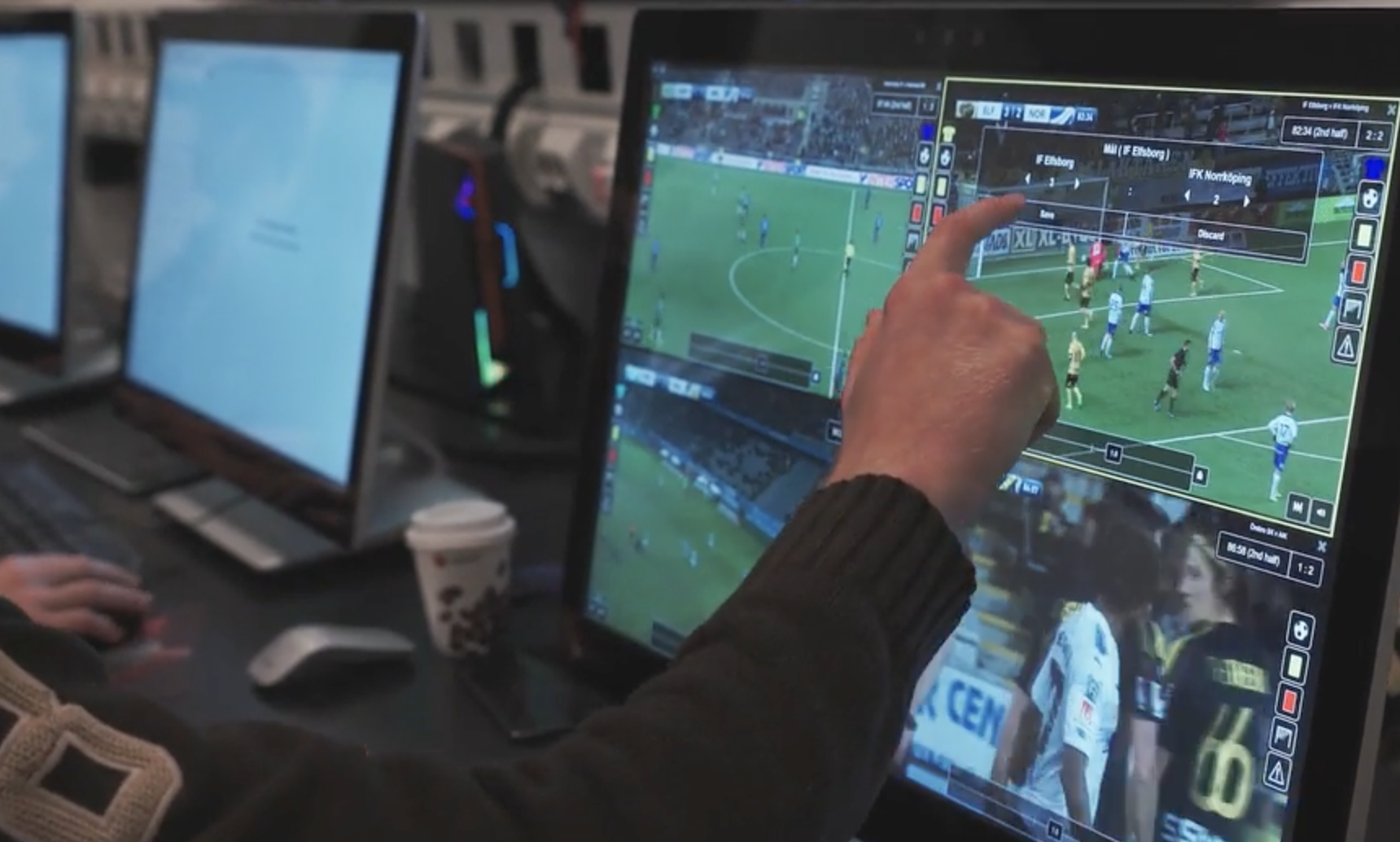}
      \caption{Detection, classification and annotation.}
      \label{figure:tagging2}
    \end{subfigure}
    \begin{subfigure}[b]{0.33\textwidth}
      \includegraphics[width=\textwidth,height=3.5cm]{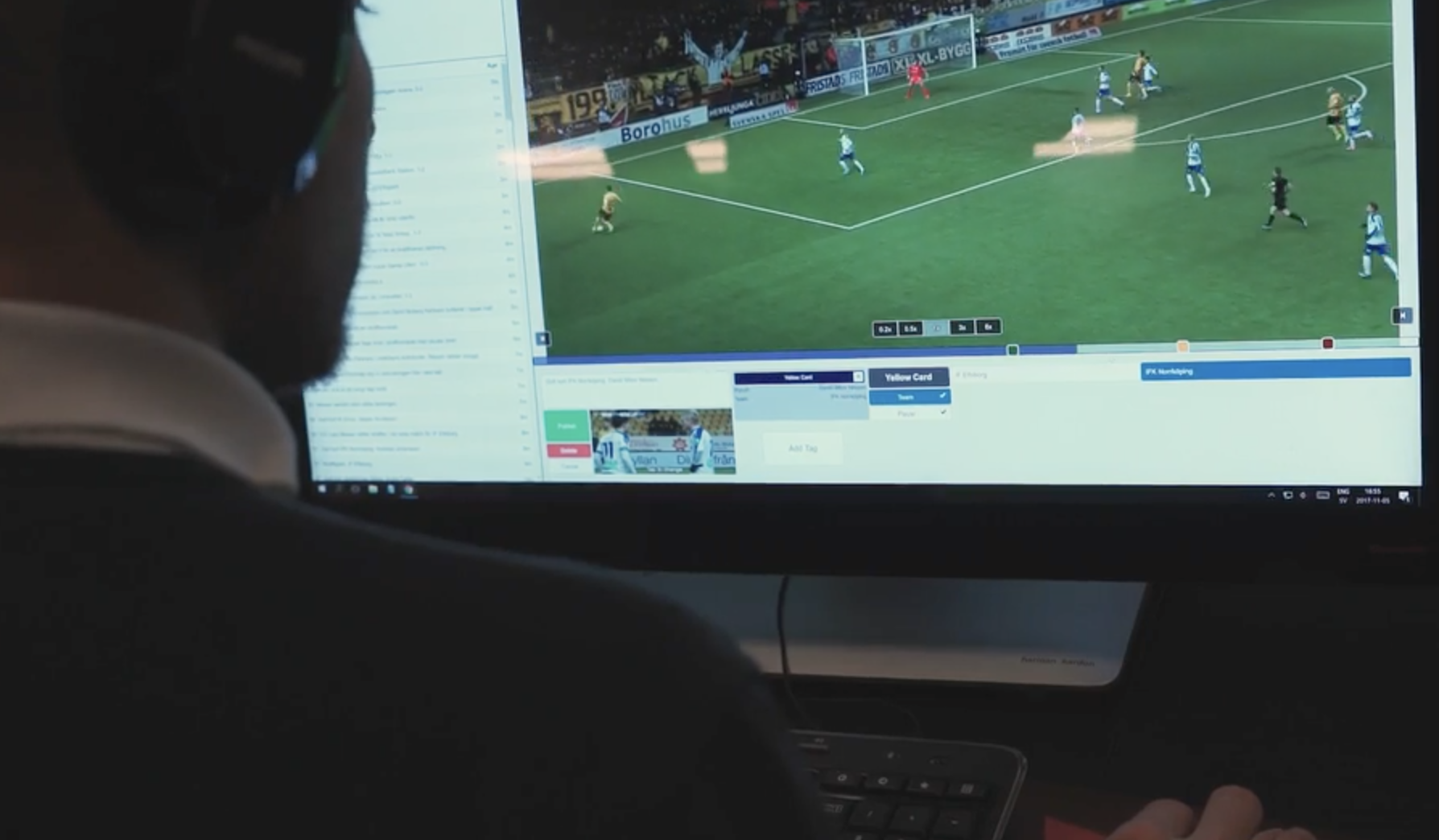}
      \caption{Clipping and thumbnail selection.}
      \label{figure:tagging3}
    \end{subfigure}
\caption{A tagging center in operation. Traditional processes are manual, cumbersome and tedious. This challenge aims to optimize and automate several of the steps involved, as described in the tasks below.}
\vspace{2mm}
\label{figure:tagging}
\end{teaserfigure}

\maketitle

\begin{figure*}
\centering
      \includegraphics[width=\textwidth]{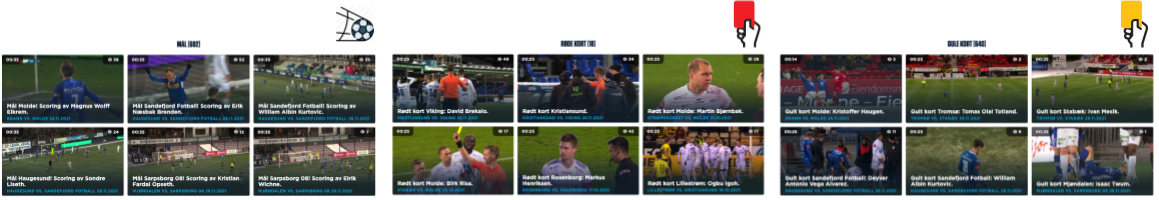}
      \caption{Highlight clip gallery examples from the Norwegian \textit{Eliteserien}. Screenshots from~\cite{web-eliteserien}.} 
      \label{figure:event-clipping-examples}
\end{figure*}

\section{Introduction}\label{section:introduction}

As one of the most popular sports, soccer had a market share of about $45\%$ of the $500$ billion global sports industry in $2020$~\cite{soccer-market-share}. Soccer broadcasting and streaming are becoming increasingly popular, as the interest in viewing videos from soccer games grows day by day. In this respect, it is important to provide game summaries and highlights, as a large percent of audiences prefer to view only the main events in a game, such as goals, cards, saves, and penalties. However, generating such summaries and event highlights requires expensive equipment and a lot of tedious, cumbersome, manual labor (Figure~\ref{figure:tagging}).

Automating and making the entire end-to-end video analysis pipeline more intelligent is seen as the ultimate goal in sports video production since it could provide fast game highlights at a much lower cost. In this context, recent developments in \ac{AI} technology have shown great potential, but state-of-the-art approaches are far from being good enough for a practical scenario that has demanding real-time requirements, as well as strict performance criteria (where at least the detection of official events such as goals and cards must be $100\%$ accurate).

Even though the event detection and classification (spotting) operation have by far received the most attention~\cite{Karpathy2014, Simonyan2014, Lin2018, Tran2018, Cioppa2019, Lin2019, Rongved2020, Rongved2021Using, Rongved2021}, it is maybe the most straightforward manual operation in an end-to-end analysis pipeline, i.e., when an event of interest occurs in a soccer game, the annotator marks the event on the timeline (Figure~\ref{figure:tagging2}). However, the full pipeline also includes various other operations which serve the overall purpose of highlight and summary generation (Figure~\ref{figure:tagging3}), some of which require careful considerations such as selecting appropriate clipping points, finding an appealing thumbnail for each clip, writing short descriptive texts per highlight, and putting highlights together in a game summary, often with a time budget in order to fit into a limited time-slot during news broadcasts.

In this challenge, we assume that event detection and classification have already been undertaken\footnote{This operation is already addressed (but still far from being solved) by the research community, and several related challenges exist, including \url{https://eval.ai/web/challenges/challenge-page/1538/overview}.}.
Therefore, the goal of this challenge is to assist the later stages of the automatization of such a production pipeline, using \ac{AI}. In particular, algorithmic approaches for event clipping, thumbnail selection, and game summarization should be developed and compared on a entirely new dataset from the Norwegian Eliteserien.

\section{Tasks}\label{section:tasks}
In this challenge, we focus on event clipping, thumbnail selection, and game summarization. Soccer games contain a large number of event types, but we focus on \textit{cards} and \textit{goals} within the context of this challenge.

\subsection{Task 1: Event Clipping}\label{section:task2}
Highlight clips are frequently used to display selected events of importance from soccer games (Figure~\ref{figure:event-clipping-examples}). When an event is detected (spotted), an associated timestamp indicates when the event happened, e.g., a tag in the video where the ball passes the goal line. However, this single annotation is not enough to generate a highlight clip that summarizes the event for viewers. Start and stop timestamps are needed to extract a highlight clip from the soccer game video (e.g., clipping the frames between x seconds before the event annotation and y seconds after the event annotation).

In the area of event clipping, the amount of existing work is limited. Koumaras et al.~\cite{Koumaras2006} presented a shot detection algorithm.
Zawbaa et al.~\cite{Zawbaa2011} implemented a more tailored algorithm to handle cuts that transitioned gradually over several frames.
Zawbaa et al.~\cite{Zawbaa2012} classified soccer video scenes as long, medium, close-up, and audience/out of field.
Several papers presented good results regarding scene classification~\cite{Xu2001, Zawbaa2012, Rafiq2020}.
As video clips can also contain replays after an event, and replay detection can help to filter out irrelevant replays, Ren et al.~\cite{Ren2005} introduced the class labels play, focus, replay, and breaks.
Detecting replays in soccer games using a logo-based approach was shown to be effective using a \ac{SVM} algorithm, but not as effective using an \ac{ANN}~\cite{Zawbaa2011, Zawbaa2012}.
Furthermore, it was shown that audio may be an important modality for finding good clipping points.
Tjondronegoro et al.~\cite{Tjondronegoro2003} used audio for a summarization method, detecting whistle sounds based on the frequency and pitch of the audio.
Raventos et al.~\cite{Raventos2015} used audio features to give an importance score to video highlights.
Some work focused on learning spatio-temporal features using various \ac{ML} approaches~\cite{Simonyan2014, Tran2015, Carreira2018}.
Chen et al.~\cite{Chen2008} used an entropy-based motion approach to address the problem of video segmentation in sports events.
More recently, Valand et al.~\cite{Valand2021, Valand2021AI} benchmarked different neural network architectures on different datasets, and presented two models that automatically find the appropriate time interval for extracting goal events.

These works indicate a potential for \ac{AI}-supported clipping of sports videos, especially in terms of extracting temporal information. However, the presented results are still limited, and most works do not directly address the actual event clipping operation (with the exception of~\cite{Valand2021, Valand2021AI}). An additional challenge for this use case is that computing should be possible to conduct with very low latency, as the production of highlight clips needs to be undertaken in real-time for practical applications.

In this task, participants are asked to identify the appropriate clipping points for selected events from a soccer game, and generate one clip for each highlight, ensuring that the highlight clip captures important scenes from the event, but also removes ``unnecessary'' parts. The submitted solution should take the video of a complete soccer game, along with a list of highlights from the game in the form of event annotations, as input. The output should be one clip per each event in the provided list of highlights. The maximum duration for a highlight clip should be $90$ seconds.

\subsection{Task 2: Thumbnail Selection}\label{section:task3}
Thumbnails capture the essence of video clips and engage viewers by providing a first impression. A good thumbnail makes a video clip more attractive to watch~\cite{Song2016}. Thus, selecting an appropriate thumbnail (e.g., by extracting a frame from the video clip itself) is very important.
Traditional solutions in the soccer domain rely on the manual or static selection of thumbnails to describe highlight clips, which display important events such as goals and cards. However, such approaches can result in the selection of sub-optimal video frames as snapshots, which degrades the overall quality of the clip as perceived by the viewers, and consequently decreases viewership. Additionally, manual processes are expensive and time consuming.

Song et al.~\cite{Song2016} presented an automatic thumbnail selection system that exploits two important characteristics commonly associated with meaningful and attractive thumbnails: high relevance to video content, and superior visual aesthetic quality.
In this respect, image quality assessment~\cite{Su2011,Kim2019} also plays an important role in thumbnail selection. Recent work demonstrates the applicability of \ac{ML}, and more specifically adversarial and reinforcement learning, in the context of thumbnail selection~\cite{Apostolidis2021}.
However, a lot of work remains to be done for the implementation of automated algorithms within the soccer domain.

\begin{figure}
  \centering
  
  \begin{subfigure}[b]{\columnwidth}
  \centering
  \includegraphics[width=0.4\textwidth]{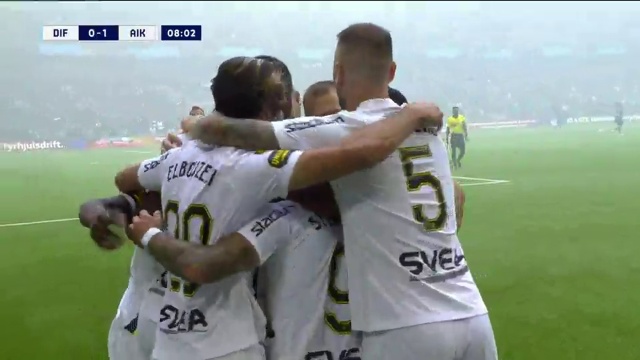}
  \includegraphics[width=0.4\textwidth]{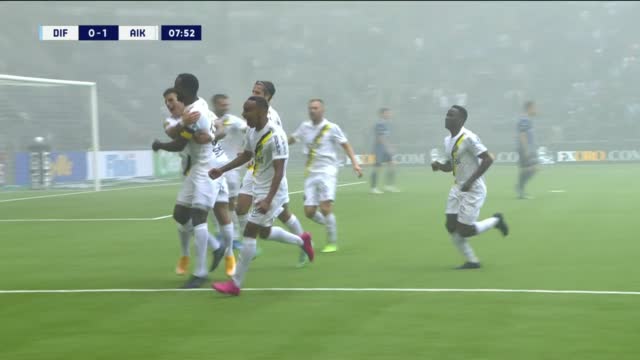}
  \vspace{-0.2cm}
  \caption{}
  \label{figure:thumbnail-selection-examples-70396}
  \end{subfigure}

  \begin{subfigure}[b]{\columnwidth}
  \centering
  \includegraphics[width=0.4\textwidth]{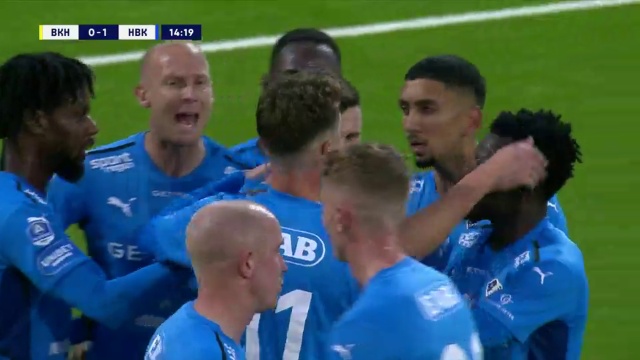}
  \includegraphics[width=0.4\textwidth]{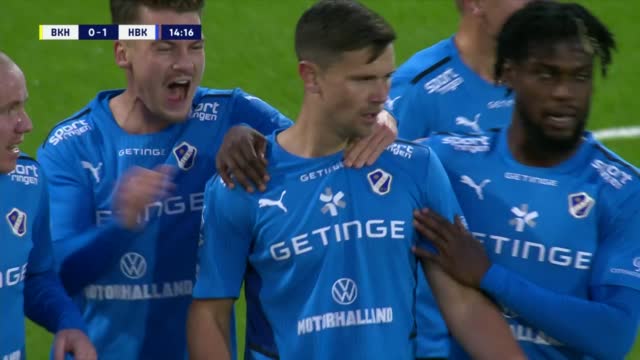}
  \vspace{-0.2cm}
  \caption{}
  \label{figure:thumbnail-selection-examples-76465}
  \end{subfigure}
  
  
  \begin{subfigure}[b]{\columnwidth}
  \centering
  \includegraphics[width=0.4\textwidth]{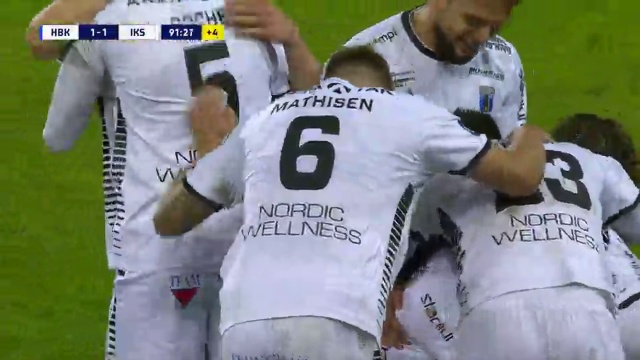}
  \includegraphics[width=0.4\textwidth]{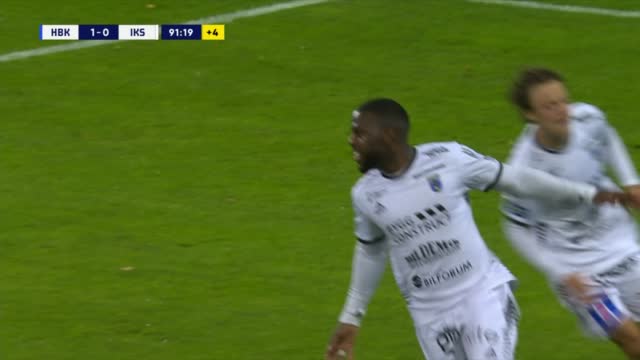}
  \vspace{-0.2cm}
  \caption{}
  \label{figure:thumbnail-selection-examples-75143}
  \end{subfigure}
  
  \begin{subfigure}[b]{\columnwidth}
  \centering
  \includegraphics[width=0.4\textwidth]{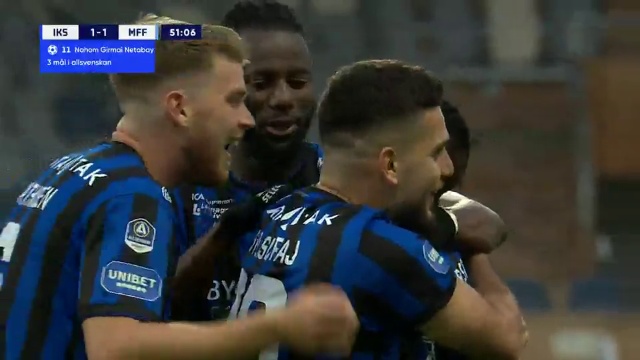}
  \includegraphics[width=0.4\textwidth]{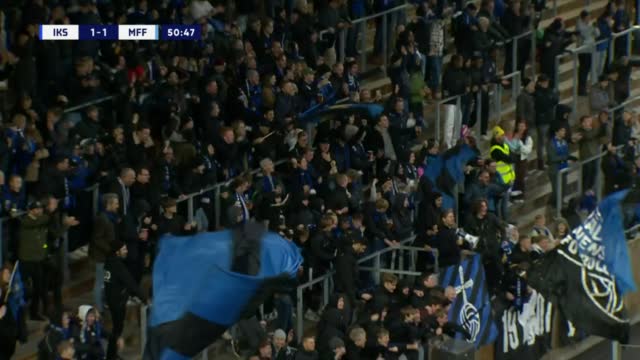}
  \vspace{-0.2cm}
  \caption{}
  \label{figure:thumbnail-selection-examples}
  \end{subfigure} 
  

\caption{Examples of alternative thumbnail candidates for the same event highlight clip.}
\label{figure:thumbnail-selection-examples}     
\end{figure}

In this task, participants are asked to identify the frame that best represents a game highlight, according to rules established by the participants themselves. The rules can be justified by references to scientific literature and industry practices. The submitted solution should take the video of a complete soccer game, along with a list of highlights from the game in the form of event annotations, as input. The output should be one image (thumbnail candidate) per each event in the provided list of highlights.

\subsection{Task 3: Game Summarization}

Soccer game summaries are of tremendous interest for multiple stakeholders including broadcasters and fans\footnote{Related challenges include \url{https://trecvid.nist.gov}}. Existing works such as~\cite{Jai-Andaloussi2014, Sanabria2019, Awad2021} consider different modalities such as video, audio, and text, but a relatively larger emphasis is put on video summaries in the broadcasting 
context.

In this task, participants are asked to generate overall game summaries for soccer games. The submitted solution should take the video of a complete soccer game, along with a list of highlights from the game in the form of event annotations, as input. The output should be a text and/or video which presents an overall summary of the game, including an adequate overview of important events, per soccer game.

\begin{itemize}
    \item \textbf{Task 3a - Text Summary:} In this subtask, participants are asked to output a text in English which serves as a summary of the soccer game, for which the maximum value for the total number of words is $100$.

    \item \textbf{Task 3b - Video Summary:} In this subtask, participants are asked to output a video (audio optional) which serves as a summary of the soccer game, for which the maximum value for the total duration of the video is $3$ minutes ($180$ seconds). How various events are ``concatenated'' into a summary is up to the participants, and using scene transition effects, as well as overlays containing detailed information (such as the names of goal scorer or booked players) are allowed.
\end{itemize}

\section{Dataset}\label{section:dataset}

\subsection{Training and Validation}\label{section:dataset-training}

An official training dataset is provided by the challenge organizers. This dataset consists of 
complete soccer game videos from the Norwegian \textit{Eliteserien}, accompanied by a list of highlights in the form of event annotations, for each game. The list of highlights includes goal annotations (Listing~\ref{lst:card-event}), card annotations (Listing~\ref{lst:goal-event}), and additional timing metadata (Listing~\ref{lst:start-end})\footnote{Each metadata list starts with the line ``Video start timestamp: <YYYY-MM-DD HH:mm:ss.ssssss''.}. 


\begin{lstlisting}[caption={Annotation structure for sample card event.},label={lst:card-event},float,floatplacement=H]
{
    '<timestamp>',
    '{
        "team": 
            { 
                "id"    : <team-id>, 
                "type"  : "team", 
                "value" : "<team-name>"
            }, 
        "action": "<yellow/red> card", 
        "player": 
            {
                "id"    : <player-id>, 
                "type"  : "player", 
                "value" : "<player-name>"
            }
    }'
}
\end{lstlisting}
\begin{lstlisting}[caption={Annotation structure for sample goal event. \\
(*) Field optional.},label={lst:goal-event},float]
{
    '<timestamp>',
    '{
        "team": 
            { 
                "id"    : <team-id>, 
                "type"  : "team", 
                "value" : "<team-name>"
            }, 
        "action": "goal", 
        "scorer": 
            {
                "id"    : <player-id>, 
                "type"  : "player", 
                "value" : "<player-name>"
            },
        "assist by":
            {
                "id"    : <player-id>,
                "type"  : "player",
                "value" : "<player-name>"
            },
        "shot type":
            {
                "type"  : "goal shot type",
                "value" : "<shot-type>"
            }
        "after set piece"(*): 
            {
                "type": "set piece", 
                "value": "penalty"
            }
    }'
}
\end{lstlisting}
\begin{lstlisting}[caption={Annotation structure for start and end timestamps.},label={lst:start-end},float]
# Start
{
    '<timestamp>',
    '{
        "phase": 
            { 
                "type"  : "phase", 
                "value" : "<1st/2nd> half"
            }, 
        "action": "start phase" 
    }'
    
}
# End
{
    '<timestamp>',
    '{
        "action": "end of game", 
    }'
}
\end{lstlisting}

In addition, prospective participants are free to use any other open dataset for training and validation purposes. In particular, interested participants are referred to the excellent and publicly available SoccerNet\footnote{https://soccer-net.org} dataset, which can be used for training and validation, as well as a transfer learning dataset for presenting additional performance results.

\subsection{Testing}\label{section:dataset-testing}

The evaluations will be undertaken using a hidden, previously unseen dataset. It will have the same format as the public training dataset provided by the challenge organizers, but will consist of completely different games.

\section{Evaluation}\label{section:evaluation}

Participants are free to develop their models in any language or platform they prefer. However, a well-documented open repository containing the source code for the proposed solution is required for each submission. Note that no data should be included within the repository itself. The hidden test dataset will be injected during evaluation, and participants can assume that the dataset will be located at \texttt{/mmsys22soccer}.

\subsection{Performance}\label{section:evaluation-performance}

As the perceived quality of highlight clips, thumbnails, and game summaries are highly subjective, the performance of the submitted solutions will be evaluated by a jury. In particular, a subjective survey will be conducted in double blind fashion with a jury consisting of unaffiliated video experts selected by the challenge organizers.
For each submitted solution for a given task, the jury members will be asked to provide an overall subjective performance score out of $100$.

\subsection{Complexity}\label{section:evaluation-complexity}

Complexity is a factor influencing how well a solution can satisfy practical real-time requirements.
The following objective metrics will be used to evaluate the submitted solutions in terms of complexity. Participants are asked to calculate the following metrics for their model and include these values in their manuscript:

\begin{itemize}
    \item \textbf{Latency:} Average runtime per sample (ms). / \textbf{Frame rate:} Average number of frames the submitted solution can analyze per second (fps).
    \item \textbf{Number of parameters:} Total number of trainable parameters in the submitted solution.
    \item \textbf{Model size:} Storage size (size on disk) of the submitted solution (MB).
\end{itemize}

\subsection {Final Score}

Aggregation of the subjective performance scores with the objective complexity scores per submission will be undertaken by the challenge organizers. For Task 3, the text (3a) and video (3b) subtasks are weighted $25\%$ and $75\%$, respectively.




\section{Conclusion and Outlook}\label{section:conclusion}
The MMSys'22 Grand Challenge on AI-based Video Production for Soccer addresses the task of automating end-to-end soccer video production systems. Such systems are used for generating event highlights and game summaries, which are operations typically requiring tedious manual labor.
An AI-based solution to replace the manual operations has the potential to both reduce human interactions and to yield better results, therefore providing a more cost efficient pipeline.
As elite soccer organizations rely on such systems, solutions presented within the context of this challenge might enable leagues to be broadcasted and/or streamed with less funding, at a cheaper price to fans.
This challenge presents three different tasks where the participants were asked to provide solutions for automatic event clipping, thumbnail selection, and game summary generation. Video and metadata from the Norwegian Eliteserien are used, and submissions will be evaluated both subjectively and objectively.
We hope that this challenge will help various stakeholders to contribute to the design of better performing systems and increase the efficiency of future video productions, not only for soccer and sports, but for video in general.

\begin{acks}
This research was partly funded by the Norwegian Research Council, project number 327717 (AI-producer). We also want to acknowledge Norsk Toppfotball (NTF), the Norwegian association for elite soccer, for making videos and metadata available for the challenge.
\end{acks}

\balance
\bibliographystyle{ACM-Reference-Format}
\bibliography{references.bib}

\end{document}